\documentclass[letterpaper]{article} 
\usepackage[preprint]{aaai2027}  
\usepackage[hyphens]{url}  
\usepackage{graphicx} 
\usepackage{natbib}  
\usepackage{caption} 
\usepackage{algorithm}
\usepackage{algorithmic}
\usepackage{newfloat}
\usepackage{listings}
\usepackage{comment}
\usepackage{amssymb}
\usepackage{pifont}
\usepackage{amsmath}
\DeclareCaptionStyle{ruled}{labelfont=normalfont,labelsep=colon,strut=off} 
\floatstyle{ruled}
\newfloat{listing}{tb}{lst}{}
\floatname{listing}{Listing}

\usepackage{booktabs}
\usepackage{multirow}

\usepackage[most]{tcolorbox}
\newtcolorbox{promptbox}[1][]{
  enhanced jigsaw,
  colback=blue!3!white,
  colframe=blue!50!black,
  coltitle=white,
  fonttitle=\bfseries\small,
  fontupper=\small,
  boxrule=0.8pt,
  arc=2pt,
  left=6pt, right=6pt, top=4pt, bottom=4pt,
  title=#1
}

\title{Diverse-Intent Multi-Turn Fashion Image Retrieval}
\author{
Mingqiang Tang\textsuperscript{\rm 1}, 
Haokun Wen\textsuperscript{\rm 2}, 
Meng Liu\textsuperscript{\rm 3},
Yupeng Hu\textsuperscript{\rm 3},
Weili Guan\textsuperscript{\rm 2, \rm4}, 
Xuemeng Song\textsuperscript{\rm 1}}
\affiliations{
\textsuperscript{\rm 1}Southern University of Science and Technology, Shenzhen, China\\
\textsuperscript{\rm 2}Harbin Institute of Technology (Shenzhen), Shenzhen, China\\
\textsuperscript{\rm 3}Shandong University, Jinan, China\\
\textsuperscript{\rm 4}Shenzhen Loop Area Institute, Shenzhen, China\\
tangmq2023@mail.sustech.edu.cn, 
haokunwen@outlook.com, 
mengliu.sdu@gmail.com,
huyupeng@sdu.edu.cn,
honeyguan@gmail.com, 
sxmustc@gmail.com}

\begin{document}

\maketitle

\begin{abstract}
Real-world fashion search involves interactive retrieval across multiple turns. However, existing multi-turn retrieval methods are built on a restrictive assumption that every interaction follows the same attribute-editing paradigm, leaving heterogeneous intent transitions unexplored. Moreover, existing approaches often rely on textification to bridge multimodal queries and visual retrieval, which may lose fine-grained visual cues.
To address these gaps, we introduce DIM-Fashion, a benchmark of 26K multi-turn sessions constructed from 13 fashion retrieval datasets across 7 tasks, featuring diverse intent transitions and rollback behaviors. We further propose FashionAM, an MLLM-VLP framework that directly aligns multimodal conversational queries with a fashion-oriented gallery embedding space, avoiding intermediate textification. Extensive experiments demonstrate the effectiveness of FashionAM over existing approaches. The dataset and code will be made publicly available upon acceptance.

\end{abstract}


\section{Introduction}
To accommodate diverse user needs, fashion image retrieval has evolved to encompass text-, image-, sketch-, and multimodal query-driven paradigms. However, most existing paradigms are studied in a single-turn setting, whereas real-world shopping is inherently interactive and evolves over multiple turns. To bridge this gap, several studies~\cite{yuan2021conversational,pal2023fashionntm,chen2025mai} have investigated multi-turn composed image retrieval (MTCIR). These methods have progressed from CNN/RNN-based architectures, such as Dialog Manager~\cite{guo2018dialog} and CFIR~\cite{yuan2021conversational}, to approaches based on vision-language pre-training (VLP), e.g., FashionNTM~\cite{pal2023fashionntm}, IRR~\cite{wei2023conversational}, and MAI~\cite{chen2025mai}, and more recently to LLM/MLLM-based methods, such as Fashion-GPT~\cite{chen2023fashion}, LLM4MS~\cite{barbany2024leveraging}, and ImageScope~\cite{luo2025imagescope}. 
Despite these advances, existing methods remain limited in both task formulation and methodology.

\begin{figure}[t]
\centering
\includegraphics[width=\columnwidth]{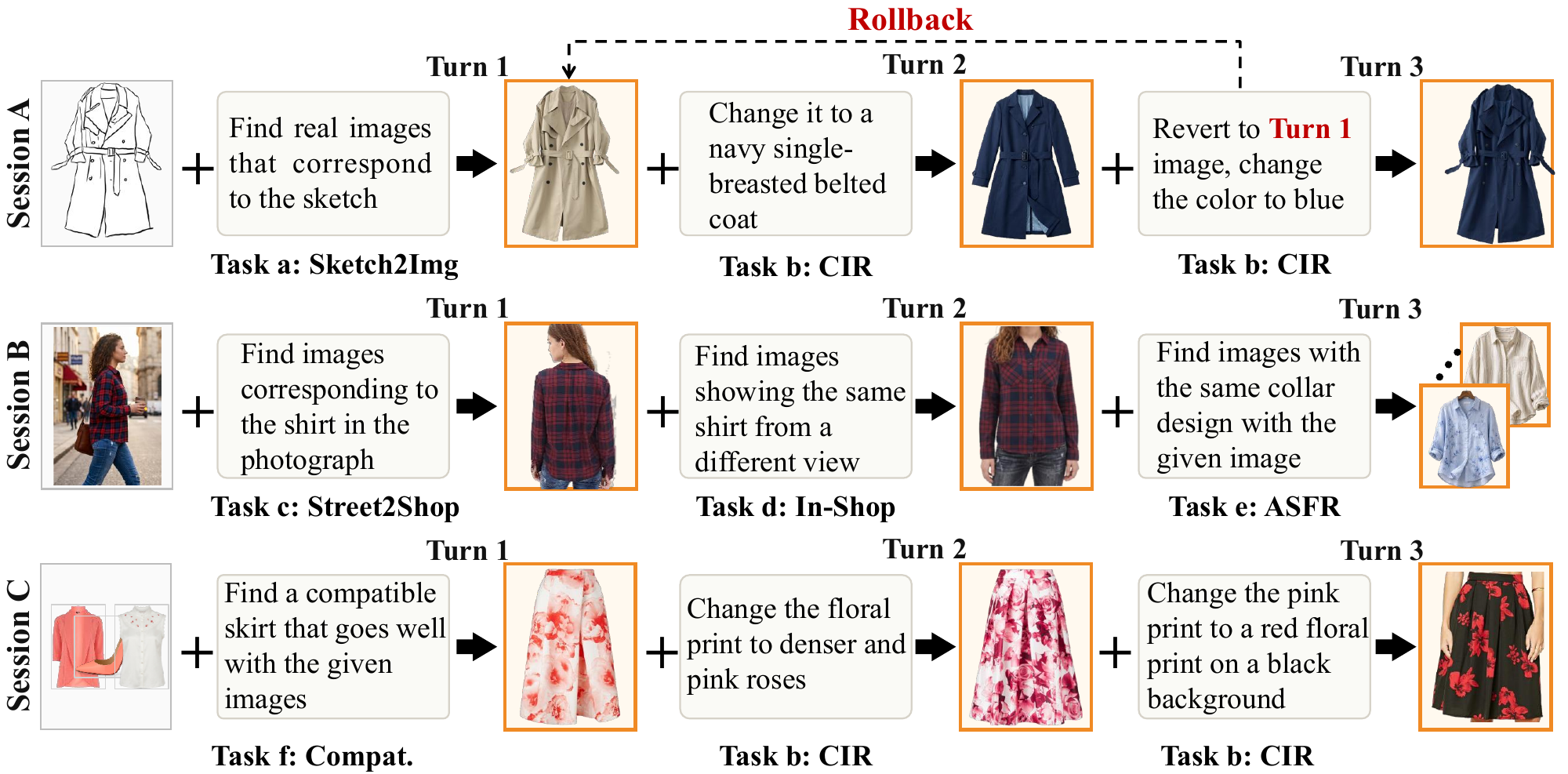}
\caption{Multi-turn retrieval sessions in DIM-Fashion. 
}
\vspace{-0.5em}

\label{fig:dim_fashion_examples}
\end{figure}

\textbf{Task level.} Existing MTCIR methods assume a homogeneous retrieval paradigm across turns, where every interaction is restricted to attribute refinement over the previous image. In realistic fashion search, however, users may express diverse retrieval intents, such as modifying an item attribute, finding the same product from another view, or searching for compatible products, and these intents may dynamically change across turns (see Figure~\ref{fig:dim_fashion_examples}). Therefore, existing formulations are insufficient for evaluating multi-turn retrieval involving both heterogeneous intents and their cross-turn transitions.

\textbf{Method level.} Although recent MTCIR methods exploit the advanced reasoning capabilities of LLMs/MLLMs, they largely follow a textification-based pipeline, where multimodal queries are first converted into textual descriptions and then used for image retrieval. Such textification introduces an information bottleneck that may discard fine-grained visual cues. This limitation becomes particularly critical in diverse-intent scenarios, where many intents are inherently image-centric, such as sketch-to-image and street-to-shop retrieval, and cannot be faithfully represented by text alone.

\begin{table*}[t]
\centering
{\small\setlength{\tabcolsep}{3pt}
\begin{tabular}{lrrcccl}
\toprule
Dataset & \#Sessions & \#Turns & Sess. Len. & Rollback & \#Tasks & Query Format \\
\midrule
MT-FashionIQ{\scriptsize~\cite{yuan2021conversational}} & 11,505 & 26,506 & 2--4 & \ding{55} & 1 & Image + Text \\
MT Shoes{\scriptsize~\cite{pal2023fashionntm}} & 4,097 & 11,346 & 2--4 & \ding{55} & 1 & Image + Text \\
FashionMT{\scriptsize~\cite{chen2025mai}} & 247,911 & 743,733 & 3 & \checkmark & 1 & Image + Text \\
\textbf{DIM-Fashion {\scriptsize(ours)}} & 26,748 & 110,841 & 2--8 & \checkmark & 7 & Text / Image+Text / Sketch+Text / Multi-image+Text\\
\bottomrule
\end{tabular}
}
\caption{Comparison with existing multi-turn fashion retrieval datasets. \emph{Session} denotes a multi-turn retrieval transaction. 
}
\label{tab:mtcir_comparison}
\end{table*}

To address the task gap, we formulate Diverse-Intent Multi-Turn Fashion Image Retrieval, where retrieval intents can dynamically shift across turns rather than being restricted to attribute refinement. Based on this formulation, we construct the DIM-Fashion benchmark through a three-step pipeline that organizes diverse fashion retrieval datasets into unified multi-turn sessions with heterogeneous search intents.
First, \textit{Cross-Dataset Bridge Pair Discovery} identifies visually similar items across heterogeneous datasets to enable cross-instance transitions. Second, \textit{Multi-Turn Session Construction} builds interaction sessions via a constrained expansion process that incorporates diverse tasks and rollback dependencies. Third, \textit{MLLM-Based Session Refinement} enriches interaction chains, while ensuring data quality through turn-level and session-level verification. As shown in Table~\ref{tab:mtcir_comparison}, DIM-Fashion surpasses prior MTCIR datasets in task diversity, query heterogeneity, and interaction length, and additionally supports retrospective rollback turns, allowing users to return to earlier retrieval states and continue the search.

To bridge the method gap, we propose FashionAM, an MLLM-VLP framework with a three-stage training pipeline, where the first two stages construct a unified fashion-oriented gallery space and the final stage directly aligns MLLM-based query representations with this space.
Stage~1 performs background-removed visual alignment, adapting the VLP image encoder to background-agnostic, item-centric representations that reduce interference from background clutter. Stage~2 refines the image encoder through lightweight item-centric vision-language alignment, where MLLM-generated captions from background-removed images are paired with raw fashion images to provide fine-grained garment-level semantic supervision. Rather than fully updating the encoder, we perform partial visual adaptation to preserve the pre-trained representation while incorporating garment-specific semantics. Stage~3 fine-tunes an MLLM with LoRA~\cite{hu2022lora} and a lightweight retrieval head to encode multi-turn multimodal contexts into query embeddings aligned with the frozen gallery space, thereby avoiding the textification bottleneck. Extensive experiments on DIM-Fashion and a public multi-turn benchmark demonstrate the effectiveness and generalization ability of FashionAM.

Our main contributions are summarized as: 1) We formulate diverse-intent multi-turn fashion image retrieval, a realistic setting where retrieval intent switches across turns rather than being restricted to attribute refinement. 2) We construct DIM-Fashion, a benchmark of 26K multi-turn sessions with heterogeneous retrieval intents. 3) We propose FashionAM, an MLLM-VLP framework that directly aligns multi-turn query representations with a fashion-oriented gallery embedding space, preserving visual information beyond text-based context encoding.

\section{Related Work}

\subsection{Multi-Turn Composed Image Retrieval}
Early MTCIR methods~\cite{guo2018dialog,yuan2021conversational} rely on CNN/RNN-based models to encode interaction history. Later methods~\cite{pal2023fashionntm,wei2023conversational,chen2025mai} adopt VLP models to learn stronger multimodal representations. Despite the advances brought by VLP, these methods remain limited in complex context reasoning and handling diverse query formats. More recently, LLM/MLLM-based approaches~\cite{chen2023fashion,barbany2024leveraging,luo2025imagescope} exploit the reasoning capability of LLMs for conversational query understanding. However, they typically involve intermediate textification introducing an information bottleneck. In contrast, FashionAM directly aligns MLLM-derived query embeddings with a VLP image embedding space, thereby improving retrieval performance.

\begin{figure*}[t]
\centering
\includegraphics[width=0.9\textwidth]{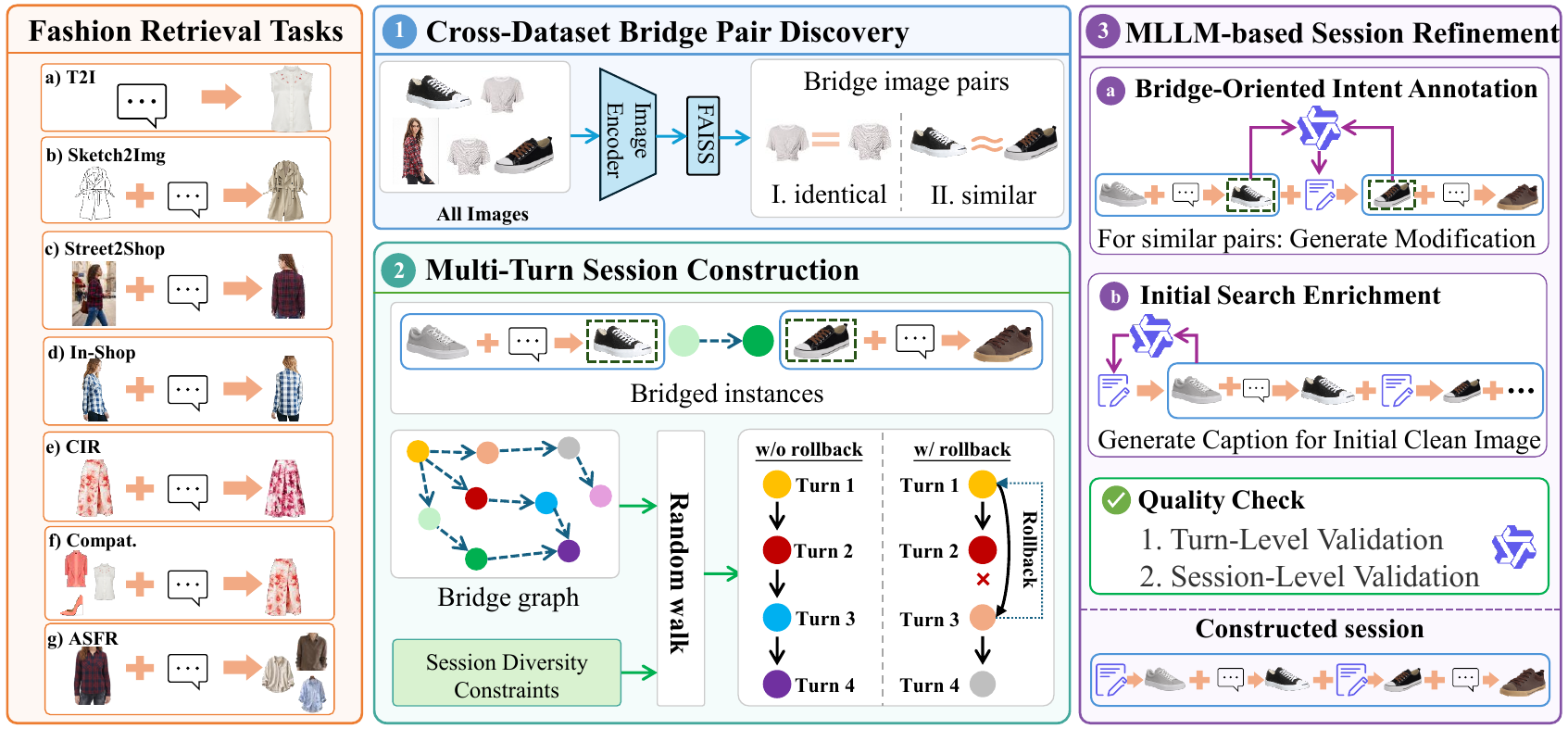}
\caption{Overview of the DIM-Fashion construction pipeline. It first discovers cross-dataset bridge image pairs, then constructs diverse multi-turn sessions through constrained random walks, and finally uses an MLLM to refine and verify the sessions.}
\vspace{-0.5em}
\label{fig:dim_fashion_construction}
\end{figure*}

\subsection{Fashion Retrieval Datasets and Benchmarks}
Existing fashion retrieval datasets are primarily designed for specific retrieval tasks, including composed image retrieval~\cite{han2017automatic,guo2018dialog,wu2021fashion}, in-shop retrieval~\cite{liu2016deepfashion}, street-to-shop retrieval~\cite{ge2019deepfashion2}, compatibility-based retrieval~\cite{han2017learning,song2017neurostylist}, attribute-specific retrieval~\cite{yu2014fine,huang2015cross,zou2019fashionai}, and sketch-based retrieval~\cite{yu2016sketch,jiang2024haifit,jiang2025arnet}. More recent benchmarks, such as U-FIRE~\cite{wen2026fashionlens}, unify multiple retrieval tasks and query formats, but remain limited to single-turn interactions. Although multi-turn benchmarks, including MT-FashionIQ~\cite{yuan2021conversational}, MT Shoes~\cite{pal2023fashionntm}, and FashionMT~\cite{chen2025mai}, enable iterative retrieval interactions, they remain restricted to a single retrieval paradigm, namely image-plus-text attribute modification.
DIM-Fashion addresses this by supporting diverse-intent multi-turn retrieval interactions. 

\section{DIM-Fashion Benchmark}

Figure~\ref{fig:dim_fashion_construction} illustrates the three-step pipeline for constructing DIM-Fashion from 13 datasets spanning seven fashion retrieval tasks: text-to-image retrieval (T2I), sketch-to-image retrieval (Sketch2Img), street-to-shop retrieval (Street2Shop), in-shop retrieval (In-Shop), composed image retrieval (CIR),  compatibility-based retrieval (Compat.), and attribute-specific fashion retrieval (ASFR). Following U-FIRE~\cite{wen2026fashionlens}, we first unify heterogeneous query formats into a representation consisting of a visual reference input and a textual instruction specifying the retrieval intent (e.g., ``\textit{finding compatible items for the given visual input}''), enabling the construction of multi-turn sessions.

\paragraph{Step 1: Cross-Dataset Bridge Pair Discovery.}
A key challenge in constructing cross-task retrieval sessions is identifying meaningful transitions across heterogeneous tasks. Ideally, identical items shared across tasks could serve as bridges between retrieval instances. However, such exact matches are extremely sparse in practice.
Thus, we relax the matching criterion to visually similar items. Specifically, we construct bridge pairs by selecting image pairs from different datasets whose cosine similarity in the VLP embedding space exceeds a predefined threshold (0.80). As a result, the discovered bridge pairs fall into two categories. Some correspond to the same fashion item appearing in different datasets, allowing retrieval turns to be directly connected across tasks. The remaining pairs consist of visually highly similar items, which provide plausible transition points and can be naturally transformed into composed image retrieval interactions during the intent enrichment process in Step 3. In practice, we use FAISS~\cite{douze2025faiss} to perform efficient nearest-neighbor search over VLP image embeddings.

\paragraph{Step 2: Multi-Turn Session Construction.}
Based on the discovered bridge image pairs, we construct multi-turn retrieval sessions through a constrained random-walk process over bridge-connected retrieval instances. A session begins from an eligible starting turn, i.e., one whose target image has at least one bridge neighbor that can serve as the reference image of another retrieval instance, so that the session can be further expanded. For subsequent turns, the reference input must be derived from a previously retrieved result to maintain coherent interaction dependencies. Considering realistic user scenarios, certain tasks (street-to-shop, sketch-to-image, and outfit compatibility retrieval) are restricted to the first turn, as their input formats (street photos, sketches, and multiple images) are more naturally suited for session-entry queries rather than follow-up queries. The remaining tasks can appear in any turn. 

Interactive search often involves rollback behaviors~\cite{chen2025mai}, where users revisit earlier results and resume retrieval from a historical state. To capture such scenarios, we introduce rollback turns from the third retrieval round onward. Specifically, when extending a session at round $t \geq 3$, the walk randomly selects an earlier round $n$ ($1 \leq n \leq t-2$) as a potential rollback point. Since rollback requires a valid transition from a historical visual state, the walk checks whether the target image of round $n$ can be connected to a CIR instance. If so, this CIR instance is appended as the new turn with a revert-and-modify instruction (e.g., ``\textit{Revert to round $n$ target image, then \ldots}''). Otherwise, the session is extended through the standard forward transition.

To encourage diverse retrieval trajectories and avoid repetitive sequences, each session is iteratively extended through bridge-connected retrieval turns subject to the following constraints: (i) a source dataset appears at most twice within a session; (ii) the same task type appears in no more than two consecutive turns; (iii) query and target images are not repeated within a session; and (iv) the first retrieval round is sampled approximately uniformly across task types for balanced coverage. The extension terminates under either of two conditions: a) when the session reaches a predefined maximum number of turns $T_\mathrm{max}$, or b) when the current target image has no bridge edge that satisfies the above constraints.

\paragraph{Step 3: MLLM-based Session Refinement.}
The resulting sessions remain limited in transition coherence and query modality coverage. First, although the bridge image pairs identified in Step 1 enable cross-dataset transitions, visually similar images may not depict the same garment, and directly replacing the reference image with the previous target image may introduce abrupt transitions. Second, real users may initiate fashion searches with text-only requests, whereas the sessions constructed in Step 2 are entirely initialized through visual bridges, limiting the coverage of text-initiated interactions.
We thus employ Qwen3.6-35B-A3B~\cite{qwen36_35b_a3b} to complete and refine the constructed sessions through two augmentations: (1) Bridge-Oriented Intent Annotation, where the MLLM identifies whether a bridge pair refers to the same garment. For visually similar but different garments, it further generates modification instructions to convert the transition into an explicit CIR turn. (2) Initial Search Enrichment, which introduces an additional text-to-image retrieval turn by generating a natural language query from initial reference images. To ensure query quality, we apply this augmentation only to item-centric images identified by the MLLM, avoiding cases where background elements may interfere with query generation.

Finally, we perform MLLM-based quality verification with Qwen3.6-35B-A3B to ensure dataset reliability. At the turn level, each retrieval turn is validated based on its task type using the reference image (if available), textual instruction, and target image. At the session level, the complete interaction history is examined for cross-turn semantic inconsistencies. Ultimately, DIM-Fashion contains 26,748 multi-turn sessions and 110,841 retrieval turns. The data are split into training, validation, and test sets in a 7:1:2 ratio, ensuring that each retrieval turn appears in only one split. Table~\ref{tab:task_distribution} reports the detailed task-level statistics. 

\begin{table}[t]
\centering
{\small\setlength{\tabcolsep}{1.2pt}
\begin{tabular}{llrrr}
\toprule
Task & Data Source & Train & Val & Test \\
\midrule
T2I & MLLM-Generated {\scriptsize(Our Step 3)} & 7,888 & 767 & 1,325 \\
\midrule
\multirow[c]{3}{*}{Sketch2Img} & ClothesV1{\scriptsize ~\cite{jiang2025arnet}} & 50 & 39 & 116 \\
& HAIFashion{\scriptsize ~\cite{jiang2024haifit}} & 89 & 33 & 106 \\
& QMUL-Shoe-V2{\scriptsize ~\cite{yu2016sketch}} & 251 & 100 & 743 \\
\midrule
Street2Shop & DeepFashion2{\scriptsize ~\cite{ge2019deepfashion2}} & 1,974 & 378 & 652 \\
\midrule
In-Shop & DeepFashion{\scriptsize ~\cite{liu2016deepfashion}} & 8,513 & 1,894 & 3,258 \\
\midrule
\multirow[c]{4}{*}{CIR} & Fashion200K{\scriptsize ~\cite{han2017automatic}} & 20,643 & 2,536 & 4,236 \\
& FashionIQ{\scriptsize ~\cite{wu2021fashion}} & 831 & 58 & 147 \\
& Shoes{\scriptsize ~\cite{guo2018dialog}} & 1,076 & 200 & 1,450 \\
& MLLM-Generated {\scriptsize(Our Step 3)}  & 19,639 & 2,572 & 5,703 \\
\midrule
\multirow[c]{2}{*}{Compat.} & FashionVC{\scriptsize ~\cite{song2017neurostylist}} & 3,940 & 414 & 641 \\
& Polyvore{\scriptsize ~\cite{han2017learning}} & 9,790 & 738 & 1,438 \\
\midrule
\multirow[c]{3}{*}{ASFR} & DARN{\scriptsize ~\cite{huang2015cross}} & 1,369 & 236 & 592 \\
& FashionAI{\scriptsize ~\cite{zou2019fashionai}} & 2,406 & 450 & 911 \\
& UT-Zappos{\scriptsize ~\cite{yu2014fine}} & 329 & 121 & 199 \\
\bottomrule
\end{tabular}
}
\caption{Task-level statistics of DIM-Fashion.} \label{tab:task_distribution}
\vspace{-0.5em}
\end{table}

\section{Method}
\label{sec:method}

\begin{figure*}[t]
\centering
\includegraphics[width=0.9\textwidth]{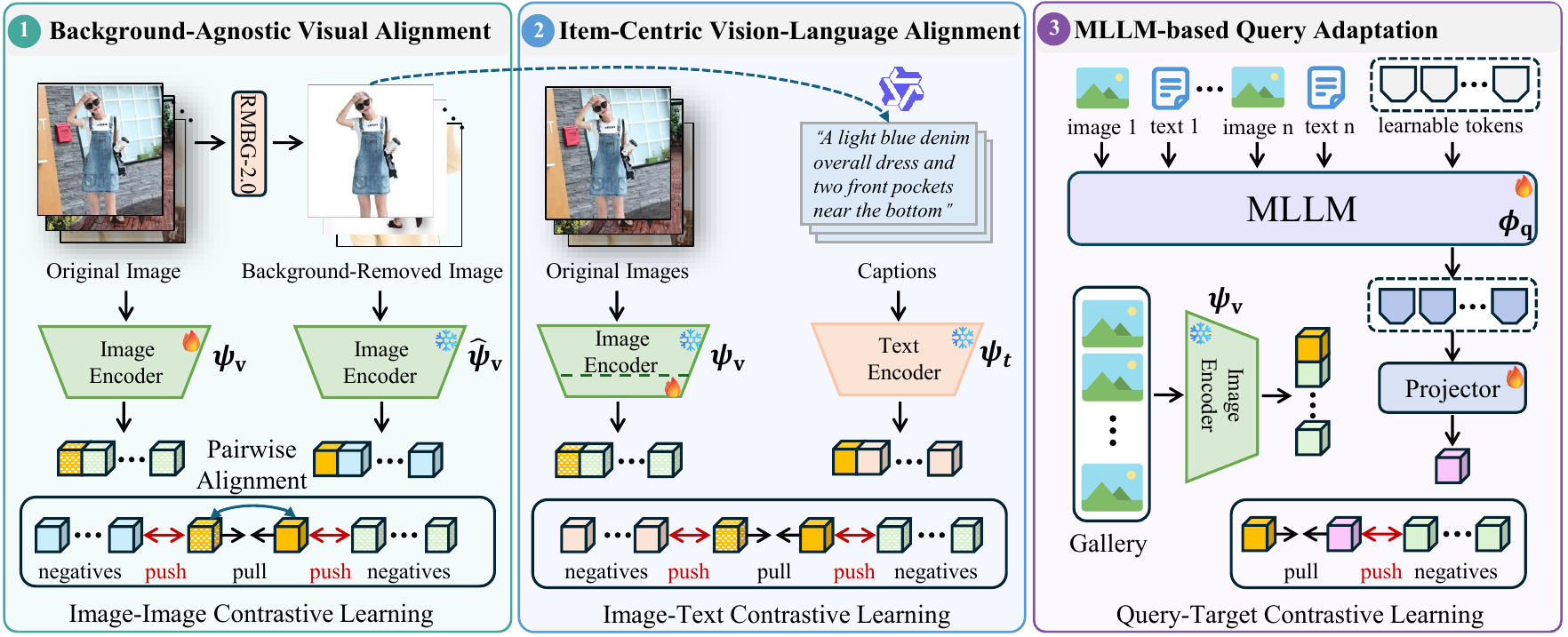}
\caption{Overview of FashionAM. Stages~1 and~2 learn a fashion-oriented gallery embedding space, while Stage~3 freezes the gallery encoder and fine-tunes an MLLM context encoder to align query representations with the learned embedding space.}
\vspace{-0.5em}
\label{fig:fashionam_framework}
\end{figure*}

\subsection{Problem Formulation}
Formally, each session consists of a sequence of $T$ retrieval turns. For turn $t$, the user query is represented as $q_t = (\mathcal{R}_t, x_t)$, where $\mathcal{R}_t$ denotes the visual reference input and $x_t$ denotes the textual instruction specifying the current retrieval intent. Depending on the search intent of turn $t$, $\mathcal{R}_t$ may be an empty set, a single image, a sketch image, or a collection of  multiple reference images. For the initial turn $t=1$, $\mathcal{R}_1$ is directly provided by the user as the starting visual cue, which can be empty for text-only initial search. For subsequent turns ($t>1$), $\mathcal{R}_t$ is obtained from the preceding interaction context. Following prior work~\cite{chen2025mai}, we adopt an oracle training protocol to mitigate error accumulation, where $\mathcal{R}_t$ is replaced with the ground-truth target image of the $(t-1)$-th turn during training.

In this work, we aim to retrieve target images from a gallery set $\mathcal{G} = \{I_1, I_2, \ldots, I_N\}$ given a multi-turn interaction context. Formally, we define the interaction context at turn $t$ as $C_t = \{q_t, \mathcal{H}_t\}$, where $q_t$ denotes the current query and $\mathcal{H}_t = \{q_i\}_{i=1}^{t-1}$ represents historical interactions. To model the heterogeneous context $C_t$, we employ an MLLM $\phi_\mathrm{q}(\cdot)$ as the query encoder. For gallery images, we use a VLP image encoder $\psi_\mathrm{v}(\cdot)$ to obtain visual representations. Overall, we aim to align the query representation produced by $\phi_\mathrm{q}(\cdot)$ with the gallery embedding space of $\psi_\mathrm{v}(\cdot)$, such that ground-truth target images $\mathcal{P}_t$ have higher similarity scores with the query context.

\subsection{Stage 1: Background-Agnostic Visual Alignment}
Raw fashion images often contain irrelevant background contexts, which may introduce visual distractions and impair retrieval performance. This stage aims to adapt the VLP image encoder $\psi_\mathrm{v}(\cdot)$ to extract background-agnostic and item-centric visual representations. As shown in Figure~\ref{fig:fashionam_framework}, we first preprocess the training images in DIM-Fashion while excluding images appearing in validation and test sessions to prevent data leakage. For each raw image $I_i$, we remove its background using RMBG-2.0~\cite{BiRefNet}, obtaining a background-removed image $\hat{I}_i$. We then align the representations of $I_i$ and $\hat{I}_i$ to adapt the gallery image encoder.

Specifically, we introduce a frozen anchor encoder $\hat{\psi}_\mathrm{v}(\cdot)$ to encode $\hat{I}_i$, providing stable item-centric visual targets during alignment. Both $\psi_\mathrm{v}(\cdot)$ and $\hat{\psi}_\mathrm{v}(\cdot)$ are initialized from the same CLIP checkpoint. Let $\mathbf{v}_i$ and $\hat{\mathbf{v}}_i$ denote the normalized features extracted from $\psi_\mathrm{v}(I_i)$ and $\hat{\psi}_\mathrm{v}(\hat{I}_i)$, respectively. We then align noisy and clean representations using two complementary objectives: a symmetric contrastive loss and a pairwise alignment loss.

\textit{Symmetric Contrastive Loss}.
 We first adopt symmetric cross-entropy contrastive loss~\cite{radford2021learning} to enforce bidirectional matching between noisy raw features and clean reference features, with a learnable temperature $\tau_1$:
\begin{equation}
\begin{aligned}
\mathcal{L}_{\mathrm{ctr}}^{\mathrm{v-\hat{v}}}
=&
-\frac{1}{B}
\sum_{i=1}^{B}
\log
\frac{\exp( \mathbf{v}_i^\top \hat{\mathbf{v}}_i/\tau_1)}
{\sum_{j=1}^{B}\exp(\mathbf{v}_i^\top \hat{\mathbf{v}}_j/\tau_1)} \\
&-\frac{1}{B}
\sum_{i=1}^{B}
\log
\frac{\exp(\hat{\mathbf{v}}_i^\top \mathbf{v}_i/\tau_1)}
{\sum_{j=1}^{B}\exp(\hat{\mathbf{v}}_i^\top \mathbf{v}_j/\tau_1)}.
\end{aligned}
\end{equation}

\textit{Pairwise Alignment Loss.} To further shrink the feature distance between matched noisy-clean pairs, we add a direct pairwise cosine alignment loss as:
\begin{equation}
\mathcal{L}_{\mathrm{cos}}^{\mathrm{v-\hat{v}}}
=
1 -
\frac{1}{B}
\sum_{i=1}^{B}
\mathbf{v}_i^\top \hat{\mathbf{v}}_i.
\end{equation}

The overall training objective for this stage is
\begin{equation}
\mathcal{L}_{\mathrm{Stage1}}^{\mathrm{v-\hat{v}}} = 
\mathcal{L}_{\mathrm{ctr}}^{\mathrm{v-\hat{v}}} + \lambda \mathcal{L}_{\mathrm{cos}}^{\mathrm{v-\hat{v}}},
\end{equation}
where $\lambda$ is the weight hyperparameter, set to $0.1$ by default.
\subsection{Stage 2: Item-Centric Vision-Language Alignment}
Although Stage 1 produces background-robust and item-centric visual representations, it does not yet establish explicit cross-modal alignment with fine-grained garment semantics. To bridge this gap and learn a unified fashion-aware multimodal embedding space, we follow prior works~\cite{chia2022contrastive,baldrati2023composed} and introduce item-centric vision-language alignment in this stage.

Existing fashion retrieval methods typically rely on dataset-provided captions for image-text alignment. However, our unified framework integrates multiple heterogeneous datasets, many of which lack paired garment descriptions. To provide consistent semantic supervision, we leverage Qwen3.6-35B-A3B to generate descriptive captions for all training images. Specifically, captions are generated from background-removed images to emphasize intrinsic garment attributes, including color, material, silhouette, pattern, and style. We then align the generated captions with the corresponding original images through symmetric image-text contrastive learning with a learnable temperature $\tau_2$,
\begin{equation}
\begin{aligned}
\mathcal{L}_{\mathrm{Stage2}}^{\mathrm{v-t}}
=&
-\frac{1}{B}
\sum_{i=1}^{B}
\log
\frac{\exp( \mathbf{v}_i^\top \mathbf{t}_i/\tau_2)}
{\sum_{j=1}^{B}\exp( \mathbf{v}_i^\top \mathbf{t}_j/\tau_2)} \\
&-\frac{1}{B}
\sum_{i=1}^{B}
\log
\frac{\exp( \mathbf{t}_i^\top \mathbf{v}_i/\tau_2)}
{\sum_{j=1}^{B}\exp( \mathbf{t}_i^\top \mathbf{v}_j/\tau_2)}.
\end{aligned}
\end{equation}
where $\mathbf{v}_i$ and $\mathbf{t}_i$ denote the image and caption features extracted by the image encoder $\psi_\mathrm{v}(\cdot)$ inherited from Stage 1 and the pre-trained VLP text encoder $\psi_\mathrm{t}(\cdot)$, respectively. 

\textbf{Partial Visual Adaptation.} 
We keep $\psi_\mathrm{t}(\cdot)$ frozen and partially fine-tune $\psi_\mathrm{v}(\cdot)$ by updating only the last two transformer layers, post-layer normalization, and visual projection. This preserves the background-robust, item-centric representations learned in Stage 1 while adapting high-level visual features to fine-grained fashion semantics.
This fine-tuning scope is analyzed in Supplementary Material.

\begin{table*}[!t]
\centering
{\small\setlength{\tabcolsep}{2.5pt}
\begin{tabular}{lrrrrrrrrrrrrrrr}
\toprule
Model
& \multicolumn{2}{c}{\textbf{Overall}}
& \multicolumn{2}{c}{T2I}
& \multicolumn{2}{c}{Sketch2Img}
& \multicolumn{2}{c}{Street2Shop}
& \multicolumn{2}{c}{In-Shop}
& \multicolumn{2}{c}{CIR}
& \multicolumn{2}{c}{Compat.}
& \multicolumn{1}{c}{ASFR} \\
\cmidrule(lr){2-3}
\cmidrule(lr){4-5}
\cmidrule(lr){6-7}
\cmidrule(lr){8-9}
\cmidrule(lr){10-11}
\cmidrule(lr){12-13}
\cmidrule(lr){14-15}
\cmidrule(lr){16-16}
& R@5 & R@8
& R@5 & R@8
& R@5 & R@8
& R@5 & R@8
& R@5 & R@8
& R@5 & R@8
& R@5 & R@8
& mAP \\
\midrule
Dialog Manager {\scriptsize~\cite{guo2018dialog}}
& 19.7 & 25.1
& 26.3 & 32.5
& 6.0 & 6.3
& 6.1 & 8.3
& 14.1 & 18.6
& 25.3 & 32.2
& 4.0 & 5.3
& 27.8 \\
CFIR{\scriptsize~\cite{yuan2021conversational}}
& 29.5 & 33.2
& 34.3 & 38.9
& 17.6 & 24.0
& 38.8 & 43.3
& 35.6 & 39.5
& 31.9 & 35.7
& 5.6 & 7.1
& 32.5 \\
\midrule
MAI{\scriptsize~\cite{chen2025mai}}
& 42.2 & 51.3
& 56.2 & 66.4
& 3.5 & 8.4
& 50.9 & 62.0
& 33.3 & 41.8
& 52.1 & 62.3
& \textbf{7.4} & \textbf{12.0}
& 37.2 \\
\midrule
OpenFlamingo{\scriptsize~\cite{awadalla2023openflamingo}}
& 20.9 & 24.7
& 73.6 & 78.2
& 0.0 & 0.0
& 0.3 & 0.3
& 10.4 & 14.5
& 24.4 & 29.3
& 0.2 & 0.3
& 33.5 \\
ImageScope{\scriptsize~\cite{luo2025imagescope}}
& 44.0 & 51.8
& \textbf{77.6} & \textbf{83.2}
& 0.6 & 4.5
& 13.8 & 17.3
& 29.0 & 37.4
& 57.6 & 67.5
& 0.4 & 0.5
& 39.9 \\
Qwen3.5-4B{\scriptsize~\cite{qwen3.5}}
& 36.9 & 44.7
& 75.8 & 81.1
& 10.9 & 17.3
& 35.0 & 43.6
& 25.7 & 33.0
& 44.4 & 54.0
& 1.2 & 1.6
& 43.7 \\
\textbf{FashionAM (ours)}
& \textbf{51.0} & \textbf{60.1}
& 59.5 & 67.0
& \textbf{20.5} & \textbf{24.5}
& \textbf{58.1} & \textbf{66.9}
& \textbf{47.1} & \textbf{53.8}
& \textbf{61.3} & \textbf{72.8}
& 6.9 & 9.7
& \textbf{54.7} \\
\bottomrule
\end{tabular}
}
\caption{Results on DIM-Fashion (\%). 
\textbf{Overall} is the average R@$K$ (Recall@$K$) over all queries from the non-ASFR tasks.}
\label{tab:dim_fashion_results}
\end{table*}

\subsection{Stage 3: MLLM-based Query Adaptation}
Through training in Stages 1 and 2, $\psi_\mathrm{v}(\cdot)$ learns retrieval-oriented visual representations that define the target embedding space. In Stage 3, we freeze $\psi_\mathrm{v}(\cdot)$ and fine-tune the MLLM context encoder $\phi_{\mathrm q}(\cdot)$ with LoRA~\cite{hu2022lora} to map diverse multimodal queries into this space.

Following the sample construction strategy of prior multi-turn retrieval work~\cite{chen2025mai}, we produce $T$ turn-level training samples for a session with $T$ retrieval turns, with the input of the $t$-th sample covering the session's initial $t$ turns. For the sample ending at turn $t$, we convert the available context $C_t=\{q_t,\mathcal{H}_t\}$ into an interleaved image-text input sequence, by sequentially arranging $\boldsymbol{R}$ and $\boldsymbol{x}$ of turns from 1 to $t$ in chronological order. We feed this ordered multimodal sequence into the MLLM, with an explicit textual instruction $\mathcal{P}_{ret}$ guiding the model to infer the target image representation from the given ordered multimodal context.
To extract a compact query representation while preserving multimodal context, following~\cite{pan2025transfer,wen2026fashionlens}, FashionAM appends $M$ learnable special tokens $\{\mathbf{s}_1,\dots,\mathbf{s}_M\}$ to the end of the MLLM input sequence. We average the hidden states of these appended tokens and apply an MLP projection to derive the final query representation:
\begin{equation}
\left\{
\begin{aligned}
&\left\{\mathbf{h}_{i,m}\right\}_{m=1}^{M} = \operatorname{MLLM}\bigl(\text{Input};\; \mathcal{P}_{ret},\; \mathbf{s}_1,\dots,\mathbf{s}_M\bigr), \\
&\mathbf{q}_i = \mathrm{Norm}\left(
\mathrm{MLP}\left(
\frac{1}{M}\sum_{m=1}^{M} \mathbf{h}_{i,m}
\right)
\right).
\end{aligned}
\right.
\end{equation}

We then use the contrastive objective for optimization with temperature $\tau_3$:
\begin{equation}
\mathcal{L}_{\mathrm{Stage3}}^{\mathrm{q-v}} =
\frac{1}{B}
\sum_{i=1}^B
-\frac{1}{|\mathcal{P}_i|}
\sum_{p \in \mathcal{P}_i}
\log
\frac{\exp(\mathbf{q}_i^\top \mathbf{v}_p/\tau_3)}
{\sum_{c \in \mathcal{C}_i}\exp(\mathbf{q}_i^\top \mathbf{v}_c/\tau_3)}
\end{equation}
where $\mathcal{P}_i$ is the set of targets for query $\mathbf{q}_i$, and $\mathcal{C}_i$ is the candidate set. $\mathbf{v}_c$ is the gallery image embedding pre-computed with the Stage-2 encoder $\psi_\mathrm{v}(\cdot)$ and cached before training.

\begin{table*}[!t]
\centering
{\small\setlength{\tabcolsep}{3pt}
\begin{tabular}{lrrrrrrrrrrrr}
\toprule
Model
& \multicolumn{3}{c}{\textbf{Overall}}
& \multicolumn{3}{c}{Dress}
& \multicolumn{3}{c}{Shirt}
& \multicolumn{3}{c}{Tops\&Tees} \\
\cmidrule(lr){2-4}
\cmidrule(lr){5-7}
\cmidrule(lr){8-10}
\cmidrule(lr){11-13}
& R@5 & R@8 & MRR
& R@5 & R@8 & MRR
& R@5 & R@8 & MRR
& R@5 & R@8 & MRR \\
\midrule
Dialog Manager$^\dagger${\scriptsize ~\cite{guo2018dialog}}
& 13.1 & 15.2 & 11.6
& 12.7 & 16.7 & 10.8
& 13.9 & 17.7 & 11.6
& 11.6 & 15.8 & 10.3 \\
CFIR$^\dagger${\scriptsize ~\cite{yuan2021conversational}}
& 30.3 & 33.4 & 26.5
& 29.8 & 33.5 & 25.6
& 30.5 & 34.1 & 27.4
& 29.4 & 33.6 & 26.1 \\
\midrule
IRR$^\dagger${\scriptsize ~\cite{wei2023conversational}}
& 25.2 & 29.2 & 19.4
& 26.8 & 31.2 & 20.6
& 25.8 & 30.4 & 19.8
& 27.1 & 31.7 & 20.9 \\
FashionNTM$^\dagger${\scriptsize ~\cite{pal2023fashionntm}}
& 45.7 & 50.4 & \textemdash
& 48.3 & 52.8 & \textemdash
& 43.8 & 48.8 & \textemdash
& 45.1 & 49.8 & \textemdash \\
MAI{\scriptsize ~\cite{chen2025mai}}
& 35.4 & 43.1 & 25.1
& 35.0 & 43.0 & 24.7
& 31.8 & 39.0 & 22.4
& 43.5 & 51.0 & 31.6 \\
\midrule
OpenFlamingo{\scriptsize ~\cite{awadalla2023openflamingo}}
& 13.2 & 16.1 & 9.9
& 10.9 & 14.8 & 8.1
& 17.9 & 21.0 & 13.5
& 22.0 & 26.3 & 16.5 \\
ImageScope{\scriptsize ~\cite{luo2025imagescope}}
& 30.8 & 36.7 & 23.6
& 22.9 & 28.4 & 16.7
& 41.6 & 47.0 & 30.4
& 44.8 & 51.6 & 33.2 \\
Qwen3.5-4B{\scriptsize ~\cite{qwen3.5}}
& 24.7 & 29.5 & 17.7
& 17.7 & 22.2 & 12.8
& 31.3 & 37.4 & 22.4
& 37.1 & 43.5 & 25.8 \\
\textbf{FashionAM (ours)}
& \textbf{56.3} & \textbf{61.7} & \textbf{44.9}
& \textbf{56.4} & \textbf{60.9} & \textbf{45.4}
& \textbf{53.7} & \textbf{59.0} & \textbf{42.7}
& \textbf{58.6} & \textbf{65.2} & \textbf{46.2} \\
\bottomrule
\end{tabular}
}
\caption{Results on MT-FashionIQ (\%). $\dagger$ denotes results reported in the original papers; others are reproduced by us.}
\label{tab:mt_fashioniq_results}
\end{table*}

\section{Experiments}

\subsection{Experimental Settings}

\paragraph{Datasets and evaluation metrics.} 
We conduct main experiments on DIM-Fashion.
Unlike existing evaluations that only consider the final turn, we adopt turn-level evaluation, as each turn forms a valid retrieval request conditioned on previous interactions. FashionAM involves two types of ground truth. All tasks except ASFR follow a single-positive setting, where each query has one target image, while ASFR specifies only target attributes and thus contains multiple target images per query. Accordingly, for non-ASFR tasks, we follow~\cite{yuan2021conversational} to evaluate queries over the full gallery and report Recall@$K$ ($K\in\{5,8\}$). For ASFR, we evaluate each query on its predefined candidate set derived from the original dataset, containing 7--10 positive and 50 negative images, and report mAP for ranking evaluation.

\paragraph{Implementation details.}
We initialize the image encoder from CLIP ViT-L/14~\cite{radford2021learning}. Stages 1 and 2 are each trained for 32 epochs with a batch size of 256, using learning rates of $1\times10^{-5}$ and $5\times10^{-6}$, respectively. In Stage 3, we adopt Qwen3.5-4B~\cite{qwen3.5} as the MLLM backbone and fine-tune it with rank-16 LoRA adapters. Based on the empirical analysis of learnable query tokens for multimodal understanding in prior work~\cite{pan2025transfer}, we set the number of learnable query suffix embeddings ($M$) to 64. Stage 3 is trained for 2 epochs using bf16 precision, with learning rates of $2\times10^{-5}$ for the LoRA parameters and $1\times10^{-4}$ for the retrieval head. To reduce GPU memory usage, we employ GradCache~\cite{gao2021scaling} for an effective batch size of 256 and use gradient checkpointing. For Stages 1 and 2, $\tau_1$ and $\tau_2$ follow CLIP's learnable temperature parameterization, with the logit scale initialized to $\log(1/0.07)$ and its exponential capped at 100, while $\tau_3$ in Stage 3 is fixed at 0.07. All experiments are conducted on 4 NVIDIA L40S GPUs.

\subsection{Main Comparison on DIM-Fashion}
For comparison, we adopt representative multi-turn retrieval methods with publicly available implementations as baselines, including the CNN/RNN-based Dialog Manager~\cite{guo2018dialog}, CFIR~\cite{yuan2021conversational}, the VLP-based MAI~\cite{chen2025mai}, and the LLM/MLLM-based ImageScope~\cite{luo2025imagescope}. To accommodate the diverse query formats in DIM-Fashion, we preserve the key architecture of each baseline while introducing necessary adaptations. For fair comparison, backbones are replaced with CLIP ViT-L/14 or Qwen3.5-4B when appropriate, while stronger backbones used in the original baselines are retained to preserve their model capacity. Details are provided in Supplementary Material. Following MAI, we also adapt general-purpose MLLMs, including OpenFlamingo~\cite{awadalla2023openflamingo} and Qwen3.5-4B, into training-free textification baselines by prompting them to generate a target image caption from the multi-turn context, which is then encoded by the CLIP ViT-L/14 text encoder to retrieve images.

Table~\ref{tab:dim_fashion_results} summarizes the overall and task-level performance on DIM-Fashion.
As shown,   
\textbf{1) FashionAM achieves the best overall performance across all models.} It outperforms VLP-based MAI by 8.8/8.8 points and MLLM-based ImageScope by 7.0/8.3 points in overall R@5/R@8, demonstrating the superiority of our three-stage training pipeline.
\textbf{2) FashionAM performs better when the query involves rich visual information.} On CIR, FashionAM surpasses ImageScope and MAI by 3.7 and 9.2 in R@5, respectively. The gains are even larger on image-centric tasks, reaching 58.1 vs. 13.8 on street-to-shop and 20.5 vs. 0.6 on sketch-to-image retrieval against ImageScope. Compared with the Qwen3.5-4B textification baseline (sharing the same backbone as FashionAM), FashionAM improves R@5 by 23.1 and 9.6 points on the two tasks, respectively, demonstrating that direct multimodal alignment better preserves visual cues than textification beyond backbone scaling.
\textbf{3) FashionAM underperforms textification-based methods on text-to-image retrieval.} The performance gap stems from the different optimization objectives of the two paradigms. The textification baselines retrieve using LLM-generated textual queries, tailored for text-only retrieval, whereas FashionAM maps all queries into a single fashion-specialized embedding space jointly optimized for heterogeneous query modalities, as required by the diverse-intent setting. This unified space may lead to lower performance on pure text-to-image matching but delivers substantial gains on image-centric tasks.

\begin{table}[t]
\centering
{\small\setlength{\tabcolsep}{4pt}
\begin{tabular}{lrrrr}
\toprule
Variant & R@1 & R@5 & R@8 & mAP \\
\midrule
w/o Stages 1 \& 2 {\scriptsize (Original CLIP)}
& 11.0 & 31.3 & 38.8 & 48.3 \\
w/o Stage 2
& 12.1 & 34.4 & 41.8 & 52.5 \\
w/o Stage 1
& 17.6 & 48.2 & 56.4 & 50.7 \\
w/o Pair-Align
& 19.0 & 48.9 & 57.3 & 53.4 \\
\midrule
FashionAM {\scriptsize (Qwen3.5-4B)} & 19.7 & 51.0 & 60.1 & 54.7 \\
FashionAM {\scriptsize (Qwen3.5-9B)}
& \textbf{21.8} & \textbf{53.4} & \textbf{62.4} & \textbf{57.1} \\
\bottomrule
\end{tabular}
}
\caption{Ablation study on DIM-Fashion (\%). R@$K$ is reported for non-ASFR tasks, while mAP is reported for ASFR.}
\label{tab:ablation}
\end{table}
\vspace{-0.5em}

\begin{figure}[t]
\centering
\includegraphics[width=0.9\columnwidth]{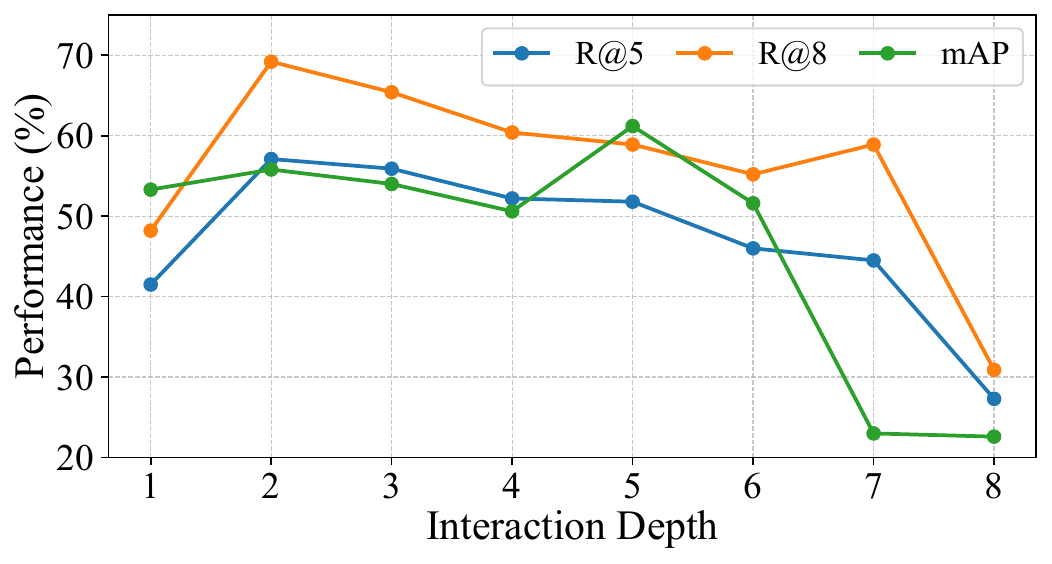}
\caption{Performance under different context lengths.}\label{fig:interaction_depth_performance}
\vspace{-0.5em}
\end{figure}

\paragraph{Performance by Context Length.}
To analyze fine-grained model behavior under different historical context lengths (i.e., numbers of turns, denoted as $L$), we visualize the corresponding performance on DIM-Fashion in Figure~\ref{fig:interaction_depth_performance}. R@$K$ metrics are computed on non-ASFR queries, while mAP is used for ASFR queries.
As shown, across all evaluation metrics, model performance initially increases and then decreases as more preceding turns are incorporated as historical context. Two major factors account for this pattern.
First, a high proportion of inherently difficult initial-stage tasks appears in single-turn samples, including sketch-to-image retrieval and outfit compatibility matching, which degrades performance at $L=1$.
Second, cues within a single round are limited, whereas incorporating additional  context introduces complementary information, which helps enrich the understanding of search intent and improves performance. However, excessively long interaction histories may introduce outdated intentions and redundant noise, which can overwhelm useful signals and degrade performance.

\subsection{Generalization on MT-FashionIQ}
To evaluate the generalization of FashionAM to standard multi-turn composed fashion retrieval, we further conduct experiments on a publicly available benchmark, MT-FashionIQ. Following its original protocol, we use the same splits, gallery definitions, evaluation setting, and metrics (Recall@5, Recall@8, and MRR) for fair comparison. Since MT-FashionIQ has existing reported results for IRR~\cite{wei2023conversational} and FashionNTM~\cite{pal2023fashionntm}, we additionally include these two methods for comprehensive comparison.

As shown in Table~\ref{tab:mt_fashioniq_results}, FashionAM achieves the best performance across all categories, demonstrating strong generalization to MTCIR. Interestingly, the VLP-based FashionNTM outperforms textification-based MLLM baselines on almost all metrics. This can be attributed to its task-specific design for multi-turn fashion retrieval, which explicitly models the accumulation and evolution of historical user preferences through dedicated memory mechanisms. This observation suggests that strong multimodal reasoning capability alone does not necessarily guarantee optimal retrieval performance, and effective modeling of retrieval-oriented representations and interaction dynamics remains crucial.

\subsection{Ablation Study}
We conduct ablation studies on both DIM-Fashion and MT-FashionIQ. The results on DIM-Fashion are presented in Table~\ref{tab:ablation}, while those on MT-FashionIQ are provided in Supplementary Material. Overall, we make the following observations. 1) Directly employing the vanilla CLIP encoder delivers the worst performance, demonstrating that generic CLIP embeddings are ill-suited for diverse fashion retrieval. 2) Removing either Stage 1 or Stage 2 harms performance, confirming they jointly refine gallery embeddings. 3) Removing Stage 2 causes larger drops in R@$K$ than removing Stage 1, highlighting the importance of fashion-semantic alignment for constructing the gallery embedding space.
4) Removing Stage 1's pairwise alignment loss hurts performance, showing that direct alignment between raw and background-erased images enhances item-centric visual embeddings.
5) Replacing Qwen3.5-4B with Qwen3.5-9B improves both R@$K$ and mAP, confirming that a stronger MLLM backbone provides better multi-turn query understanding.

\section{Conclusion}
In this paper, we study diverse-intent multi-turn fashion image retrieval, a new problem setting that enables diverse retrieval intents and better reflects realistic shopping scenarios. To support this setting, we introduce DIM-Fashion, a benchmark with 26,748 multi-turn sessions covering 7 retrieval tasks, heterogeneous query formats, and rollback-style dependencies. We further propose FashionAM, an MLLM-VLP framework with a three-stage fine-tuning pipeline. Extensive experiments on DIM-Fashion and MT-FashionIQ demonstrate its superiority over existing methods.

\bibliography{aaai2027}


\clearpage
\appendix

\twocolumn[
\begin{center}
{\LARGE\bfseries Diverse-Intent Multi-Turn Fashion Image Retrieval\par}
\vspace{1em}
{\Large Supplementary Material\par}
\end{center}
\vspace{8em}
]

This supplementary document is organized into five parts. Appendix~\ref{sec:sup_dataset} reports additional DIM-Fashion statistics; Appendix~\ref{sec:sup_construction} describes additional construction details; Appendix~\ref{sec:sup_model} presents the prompt setting for Stage~3 (MLLM-based query adaptation) training; Appendix~\ref{sec:sup_experiment_setting} details the adaptation of representative baselines to DIM-Fashion; and Appendix~\ref{sec:sup_experiments} reports additional experimental results.

\section{Additional DIM-Fashion Statistics}
\label{sec:sup_dataset}

\subsection{Source Image Distribution}
Table~\ref{tab:sup_source_datasets} lists the source image distribution (i.e., the number of unique images retained from each source dataset) in our benchmark, complementing the turn-count statistics reported in the main paper.

\begin{center}
{\setlength{\tabcolsep}{8pt}
\begin{tabular}{ccr}
\toprule
Task & Source Dataset & \#Images \\
\midrule
\multirow[c]{3}{*}{Sketch2Img} & ClothesV1 & 68 \\
 & HAIFashion & 150 \\
 & QMUL-Shoe-V2 & 223 \\
\midrule
Street2Shop & DeepFashion2 & 4,953 \\
\midrule
In-Shop & DeepFashion & 2,706 \\
\midrule
\multirow[c]{3}{*}{CIR} & Fashion200K & 8,661 \\
 & FashionIQ & 713 \\
 & Shoes & 909 \\
\midrule
\multirow[c]{2}{*}{Compat.} & FashionVC & 2,952 \\
 & Polyvore & 8,186 \\
\midrule
\multirow[c]{3}{*}{ASFR} & DARN & 2,306 \\
 & FashionAI & 3,840 \\
 & UT-Zappos & 538 \\
\midrule
Total & 13 datasets & 36,205 \\
\bottomrule
\end{tabular}
}
\captionof{table}{The 13 source datasets integrated into DIM-Fashion, grouped by retrieval task, with the number of unique images retained from each.}
\label{tab:sup_source_datasets}
\end{center}

\subsection{Session-Length Distribution}
\label{sec:sup_statistics}
Table~\ref{tab:sup_chain_length} reports the session-length distribution for each split. Most sessions contain 3--6 turns, with 3-turn sessions being the most frequent.

\begin{center}
{\setlength{\tabcolsep}{5pt}
\begin{tabular}{crrrr}
\toprule
Session Length & Train & Val & Test & Total \\
\midrule
2 & 1,780 & 270 & 513 & 2,563 \\
3 & 5,057 & 946 & 1,666 & 7,669 \\
4 & 4,796 & 650 & 1,469 & 6,915 \\
5 & 3,323 & 489 & 925 & 4,737 \\
6 & 2,509 & 161 & 509 & 3,179 \\
7 & 868 & 125 & 198 & 1,191 \\
8 & 391 & 34 & 69 & 494 \\
\midrule
Total & 18,724 & 2,675 & 5,349 & 26,748 \\
\bottomrule
\end{tabular}
}
\captionof{table}{Distribution of session lengths across different splits.} \label{tab:sup_chain_length}
\end{center}

\section{DIM-Fashion Construction Details}
\label{sec:sup_construction}

\subsection{Visual Similarity Threshold Selection for Step 1}
Table~\ref{tab:sup_bridge_threshold} reports the number of cross-dataset bridge edges and the number of images with at least one bridge under different similarity thresholds.
A lower threshold typically yields dense but noisy connections. Manual inspection reveals that many low-threshold matches are caused by superficial visual similarities, such as shared backgrounds, poses, or shooting styles, rather than true semantic correspondence, whereas a higher threshold yields more reliable but sparser bridges, making it more difficult to construct long sessions. The adopted threshold $\theta{=}0.80$ preserves over $1.2$M cross-dataset edges and $172$k bridged images while keeping the false-match rate low in our manual inspection.

\begin{center}
{\setlength{\tabcolsep}{6pt}
\begin{tabular}{ccc}
\toprule
$\theta$ & Cross-dataset edges & Images with $\geq$1 bridge \\
\midrule
0.70 & 2,613,057 & 294,881 \\
0.75 & 2,010,949 & 251,946 \\
\textbf{0.80} & \textbf{1,215,121} & \textbf{172,400} \\
0.85 & 422,890 & 76,241 \\
0.90 & 45,041 & 13,068 \\
0.95 & 426 & 354 \\
\bottomrule
\end{tabular}
}
\captionof{table}{Number of cross-dataset bridge edges and bridged images under different similarity thresholds.}
\label{tab:sup_bridge_threshold}
\end{center}

\subsection{Algorithms of Steps 1 \& 2}
\label{sec:sup_construction_algorithm}
The cross-dataset bridge discovery and multi-turn session construction procedures are summarized in Algorithms~\ref{alg:bridge} and~\ref{alg:assembly}, respectively. We represent each single-turn retrieval instance as a unified tuple $s=(q,x,y,\mathrm{task},d)$, where $q$ denotes the query input (a single image, a sketch, multiple images, or empty), $x$ the textual instruction, $y$ the target image, $\mathrm{task}\in\{\text{CIR},\text{T2I},\dots\}$ the retrieval task, and $d$ the source dataset.

\begin{algorithm}[t]
\caption{Cross-Dataset Bridge Pair Discovery (Step 1)}
\label{alg:bridge}
\begin{algorithmic}[1]
\REQUIRE Image pool $\mathcal{I}$ with dataset label $d(\cdot)$; VLP image encoder $\psi$; similarity threshold $\theta$.
\STATE Extract and $\ell_2$-normalize features $f_i \gets \psi(I_i)$ for all $I_i \in \mathcal{I}$.
\STATE Build a FAISS inner-product index over $\{f_i\}$.
\FORALL{$I_i \in \mathcal{I}$}
    \STATE $\mathcal{B}(I_i) \gets \{I_j \in \mathcal{I} \mid d(I_j) \neq d(I_i),\ f_i^\top f_j \ge \theta\}$ via FAISS range search.
\ENDFOR
\ENSURE Symmetric cross-dataset bridge index $\mathcal{B}$.
\end{algorithmic}
\end{algorithm}

\begin{algorithm}[t]
\caption{Multi-Turn Session Construction (Step 2)}
\label{alg:assembly}
\begin{algorithmic}[1]
\REQUIRE Bridge index $\mathcal{B}$; single-turn instances $\mathcal{S}$; session constraints described in dataset-construction Step~2; maximum length $T_{\max}$; target number of sessions $N$.
\STATE $\mathcal{D} \gets \emptyset$.
\REPEAT
    \STATE Sample a first-turn instance $s_1$ that can be extended to a valid second turn, approximately uniformly across task types.
    \STATE Initialize $\textit{session} \gets [s_1]$ and $t \gets 2$.
    \WHILE{$t \le T_{\max}$}
        \IF{$t \ge 3$}
            \STATE Sample $n \in \{1,\dots,t{-}2\}$.
            \IF{$y_n$ connects through $\mathcal{B}$ to a CIR instance $s^{\mathrm{rb}}$ satisfying the session constraints}
                \STATE Use $y_n$ as its reference and rewrite the instruction for rollback.
                \STATE Append $s^{\mathrm{rb}}$ to $\textit{session}$.
                \STATE $t \gets t{+}1$.
                \STATE \textbf{continue}.
            \ENDIF
        \ENDIF
        \STATE $\mathcal{C} \gets$ candidates connected from $y_{t-1}$ through $\mathcal{B}$ and satisfying the session constraints.
        \STATE \textbf{break} if $\mathcal{C} = \emptyset$.
        \STATE Sample $s \in \mathcal{C}$.
        \STATE Use $y_{t-1}$ as the reference of $s$.
        \STATE Append $s$ to $\textit{session}$.
        \STATE $t \gets t{+}1$.
    \ENDWHILE
    \STATE Add $\textit{session}$ to $\mathcal{D}$.
\UNTIL{$|\mathcal{D}| = N$}
\ENSURE Multi-turn session set $\mathcal{D}$.
\end{algorithmic}
\end{algorithm}

\subsection{Prompts for Quality Verification in Step 3}
Following the main paper, we perform MLLM-based quality verification at two granularities.
\emph{Turn-level validation} verifies each retrieval turn according to its task type, using the reference image(s) (if available), the textual instruction, and the target image.
\emph{Session-level validation} feeds the complete interaction history to the MLLM and examines it for semantic inconsistencies across turns.
A session is discarded as low quality if the turn-level check flags any turn as ambiguous or weakly grounded, or if the session-level check detects a cross-turn inconsistency.
The turn- and session-level checks are summarized in Prompt~A and Prompt~B, respectively. After both MLLM-based verification steps, a total of 21,218 low-quality sessions are filtered out, leaving the 26,748 sessions in the released benchmark.

\subsection{Retrieval-Signature-Aware Data Split}
To prevent the same retrieval instance from appearing in multiple splits, we compute a retrieval signature $(r, \tilde{x}, y)$ for every turn, where $r$ is the reference image, $\tilde{x}$ is the whitespace- and case-normalized instruction, and $y$ is the target image. Sessions sharing any signature are merged into the same connected component via union-find, and each component is assigned entirely to one split.
This produces 6,477 connected components (4,797 of which are singletons) from 110,841 retrieval signatures (63,049 unique).
The split is defined at the retrieval-relation level because the same image may legitimately participate in different retrieval contexts. To prevent transductive adaptation to evaluation images, Stages~1 and~2 use only training-exclusive DIM-Fashion images, excluding any image referenced by a validation or test session.

\begin{promptbox}[Prompt A --- Turn-Level Validation]
\small\ttfamily
For each retrieval turn, you are given the reference image(s) (if any), the textual instruction, and the target image. Judge whether the target satisfies the relationship required by its task type. \\

For example, for compatibility-based retrieval judge whether the target is broadly compatible with the query item(s), and for ASFR judge whether the target shares the requested attribute.\\

Judge the main clothing item only, ignoring background, person identity, pose, accessories, and image quality. Return is\_reasonable="yes" if the relationship is broadly satisfied, and "no" if it is clearly unreasonable, mismatched, or nonsensical.\\
Return JSON only: \{"is\_reasonable":"yes|no"\}.
\end{promptbox}

\begin{promptbox}[Prompt B --- Session-Level Validation]
\small\ttfamily
Given the complete interaction history of one session, judge whether the session is semantically self-consistent across turns. The session should read as one coherent multi-turn retrieval rather than unrelated turns concatenated together, with no turn's target image or instruction contradicting another.\\[0.3em]
Return is\_reasonable="yes" if the session is consistent, and "no" if any turn contradicts another or the session is incoherent.\\
Return JSON only: \{"is\_reasonable":"yes|no"\}.
\end{promptbox}

\begin{promptbox}[Prompt C --- Stage-3 Multi-Turn Retrieval Context]
\textbf{Chat message template}

{\ttfamily\raggedright
History turn 1 query image: <image 1>\\
History turn 1 request: <text 1>\\
History turn 1 target image (also History turn 2 query image): <image 2>\\
History turn 2 request: <text 2>\\
\dots\\
History turn k-1 request: <text k-1>\\
History turn k-1 target image (also Current turn query image): <image k>\\
Retrieval instruction
}

\medskip
\noindent\textbf{Retrieval instruction}

{\ttfamily\raggedright
You are a fashion image retrieval assistant. Infer the desired target image from the ordered multimodal context. For non-first turns, use the previous-turn target image as the default visual reference. If the current request reverts to an earlier round, locate that round's target image in the history and use it instead. Follow the request to decide which visual aspects should be preserved, changed, matched, or ignored. For modification requests, apply the requested changes. For same-attribute requests, match the requested attribute while ignoring unrelated details. For compatibility requests, retrieve the requested item type that fits the provided context.\\
Current retrieval request:\\
<current turn text>\\
Use the current request as the main instruction and represent the desired target fashion image for retrieval.
}
\end{promptbox}
\section{Prompt Setting for Stage-3 Training}
\label{sec:sup_model}
During Stage~3, each turn is formatted as a user message containing the available history images, history requests, history target images, the current query image(s), and the current retrieval request.
The ordered context is assembled as interleaved text and image segments.
If the first turn provides an explicit query image, it is sent before the current retrieval request with the label ``Current turn query image 1:''.
From the second turn onward, the previous target image is treated as the current query image.
To avoid repeatedly consuming context with the same image, FashionAM sends this image only once and marks its dual role in the text label, e.g., ``History turn $k$ target image (also Current turn query image)''.
Here, the current query image denotes the latest visual state explicitly provided to the model and serves as the default reference for a standard forward turn. For a rollback turn, the effective reference is instead the earlier historical target specified by the current request. FashionAM must locate that image from the ordered multimodal history rather than receiving it again as a separately marked query image.
Thus, the MLLM observes an ordered multimodal trajectory consisting of history query images, turn-level requests, and history target images, followed by the current retrieval request.

The complete message template and retrieval instruction are shown in Prompt~C. Following common multimodal-prompt notation, angle-bracketed fields denote inputs inserted at that position, e.g., \texttt{<image>} and \texttt{<current turn text>}.
This message template is followed by $M$ learnable query embeddings appended as a suffix. Their final hidden states are pooled and projected into the frozen Stage-2 image embedding space.
The training objective is to make the resulting query embedding close to the embedding of the current turn's target image.

\section{Detailed Experimental Settings}
\label{sec:sup_experiment_setting}
To ensure a fair comparison, every trainable baseline is trained on the identical DIM-Fashion training split using supervision from each retrieval turn, while OpenFlamingo, ImageScope, and Qwen3.5-4B remain training-free and are applied zero-shot. At each turn, all history-aware baselines are provided with the same multi-turn context as FashionAM. For each baseline, we preserve its core architecture and introduce only the necessary modifications for comparison.

\paragraph{Dialog Manager.} We retain the original GRU-based dialog tracker for recurrently aggregating multi-turn information. For a fair comparison, we replace its original learned image representations and bigram-CNN-based text representations with features extracted by the frozen CLIP ViT-L/14 image and text encoders. At each turn, the image and text features are additively fused and passed to the GRU tracker.

\paragraph{CFIR.} We retain CFIR’s self-attention text encoder, TIRG feature composer, and GRU-based multi-turn aggregation. For a fair comparison under a unified visual backbone, we replace its original ResNet-152 image encoder with CLIP ViT-L/14.

\paragraph{MAI.} We retain MAI's BLIP-2 Q-Former-based iterative aggregation and token-pruning mechanism, where learnable query tokens are updated at each turn to maintain the accumulated multi-turn state. The frozen EVA-CLIP-G visual encoder is kept unchanged to preserve MAI's original visual representation capability. Notably, EVA-CLIP-G is a larger-scale vision-language backbone than the CLIP ViT-L/14 encoder used by FashionAM ($\sim$1B vs. $\sim$300M parameters). The Q-Former, projection heads, and learnable query tokens are trained on the DIM-Fashion training split under MAI's original batch-based classification objective with turn-level supervision.

\paragraph{OpenFlamingo.} We use the official OpenFlamingo-9B checkpoint (an MPT-7B language model with CLIP ViT-L/14 and gated cross-attention every four layers). Following MAI, we apply OpenFlamingo to DIM-Fashion in a zero-shot manner, without task-specific fine-tuning. For each query, it autoregressively generates a textual target description conditioned on the interleaved reference-image and modification-text history. The generated description is encoded by the CLIP ViT-L/14 text encoder and matched against the gallery image embeddings from the same CLIP visual encoder by cosine similarity.

\paragraph{ImageScope.}
ImageScope is a training-free three-stage reason-and-retrieve, verify-and-rerank, and evaluate pipeline that combines an LLaVA-v1.6-Vicuna-7B captioner, an LLaMA-3-8B-Instruct reasoner, a PaliGemma-3B-mix-224 verifier, an InternVL2-8B evaluator, and a CLIP ViT-L/14 retriever. For a controlled backbone comparison with FashionAM, we replace ImageScope's original LLaMA-3-8B-Instruct reasoner with the same Qwen3.5-4B base backbone used by FashionAM's MLLM query encoder, while retaining the remaining pretrained components and the original pipeline logic. 

\paragraph{Qwen3.5-4B.} We also adapt Qwen3.5-4B as a training-free textification baseline. Using the same backbone as FashionAM enables a controlled comparison that isolates the contribution of retrieval-oriented alignment from backbone choice. Given the ordered reference images and retrieval requests in the multi-turn context, Qwen3.5-4B greedily generates a one-sentence description of the desired target item. The generated description is encoded by the CLIP ViT-L/14 text encoder, and retrieval is performed by cosine similarity against gallery embeddings produced by the corresponding visual encoder.

\section{Additional Experimental Results}
\label{sec:sup_experiments}

\subsection{On Stage-2 Fine-Tuning Scope}
\label{sec:sup_stage2_scope}
We further analyze which modules should be fine-tuned during Stage~2.
All variants are initialized from the Stage-1 image encoder and trained with the same item-centric vision-language alignment objective, while differing only in the fine-tuning scope.
Table~\ref{tab:sup_stage2_scope} reports the results on DIM-Fashion.

\begin{center}
{\small\setlength{\tabcolsep}{4pt}
\begin{tabular}{@{}ccccc@{}}
\toprule
Stage-2 Fine-Tuning Scope & R@1 & R@5 & R@8 & mAP \\
\midrule
Last 2 img-enc layers + txt-enc & 19.0 & 47.0 & 56.4 & 51.0 \\
All img-enc layers & 18.9 & 47.0 & 55.1 & 50.3 \\
Last 1 img-enc layer & 18.8 & 49.8 & 59.4 & 54.3 \\
Last 2 img-enc layers & 19.7 & 51.0 & \textbf{60.1} & \textbf{54.7} \\
Last 3 img-enc layers & \textbf{20.0} & \textbf{51.8} & 60.0 & 54.3 \\
\bottomrule
\end{tabular}
}
\captionof{table}{Ablation study on the fine-tuning scope of Stage~2 on DIM-Fashion (\%).}
\label{tab:sup_stage2_scope}
\end{center}

The results show that Stage~2 benefits most from lightweight adaptation of high-level visual features. Fine-tuning only the last three image-encoder layers achieves the best R@1 (20.0) and R@5 (51.8), while fine-tuning the last two image-encoder layers yields the best R@8 (60.1) and mAP (54.7). Since the performance differences are marginal, we adopt the latter as the default setting, as it offers a better trade-off between retrieval performance and computational efficiency by updating fewer parameters. In contrast, updating the entire image encoder or jointly optimizing the text encoder consistently degrades performance. We attribute this degradation to excessive representation drift: full visual fine-tuning may weaken the background-robust, item-centric representations learned in Stage~1, while updating the text encoder may disrupt the pretrained cross-modal semantic space that serves as a stable alignment anchor.

\subsection{Additional Ablation on MT-FashionIQ}
\label{sec:sup_mtfiq_ablation}
Table~\ref{tab:sup_mtfiq_ablation} reports the per-category ablation on MT-FashionIQ in terms of R@5, complementing the DIM-Fashion ablation in the main paper on a standard sequential attribute-modification benchmark. The results exhibit consistent trends with those observed on DIM-Fashion.

\begin{center}
{\setlength{\tabcolsep}{2pt}
\begin{tabular}{lrrrr}
\toprule
Variant
& Overall & Dress & Shirt & Tops\&Tees \\
\midrule
w/o Stages 1 \& 2
& 47.0 & 44.3 & 47.6 & 50.3 \\
w/o Stage 2
& 48.9 & 45.2 & 49.3 & 53.5 \\
w/o Stage 1
& 55.2 & 54.5 & 52.9 & 58.3 \\
w/o Pair-Align
& 55.7 & 56.6 & 53.3 & 56.7 \\
\midrule
FashionAM {\small (Qwen3.5-4B)}
& 56.3 & 56.4 & 53.7 & 58.6 \\
FashionAM {\small (Qwen3.5-9B)}
& \textbf{57.2} & \textbf{56.6} & \textbf{55.5} & \textbf{59.5} \\
\bottomrule
\end{tabular}
}
\captionof{table}{Ablation study on MT-FashionIQ (R@5, \%). \textit{w/o Pair-Align} removes the pairwise cosine alignment loss in Stage~1.}
\label{tab:sup_mtfiq_ablation}
\end{center}

\subsection{Case Study}
\label{sec:sup_case_studies}
\begin{figure*}[!t]
\centering
\noindent\makebox[\textwidth][c]{%
  \begin{minipage}[c]{0.04\textwidth}
    \raggedleft\textbf{(a)}
  \end{minipage}\hspace{0.01\textwidth}%
  \begin{minipage}[c]{0.87\textwidth}
    \centering\includegraphics[page=1,width=\linewidth]{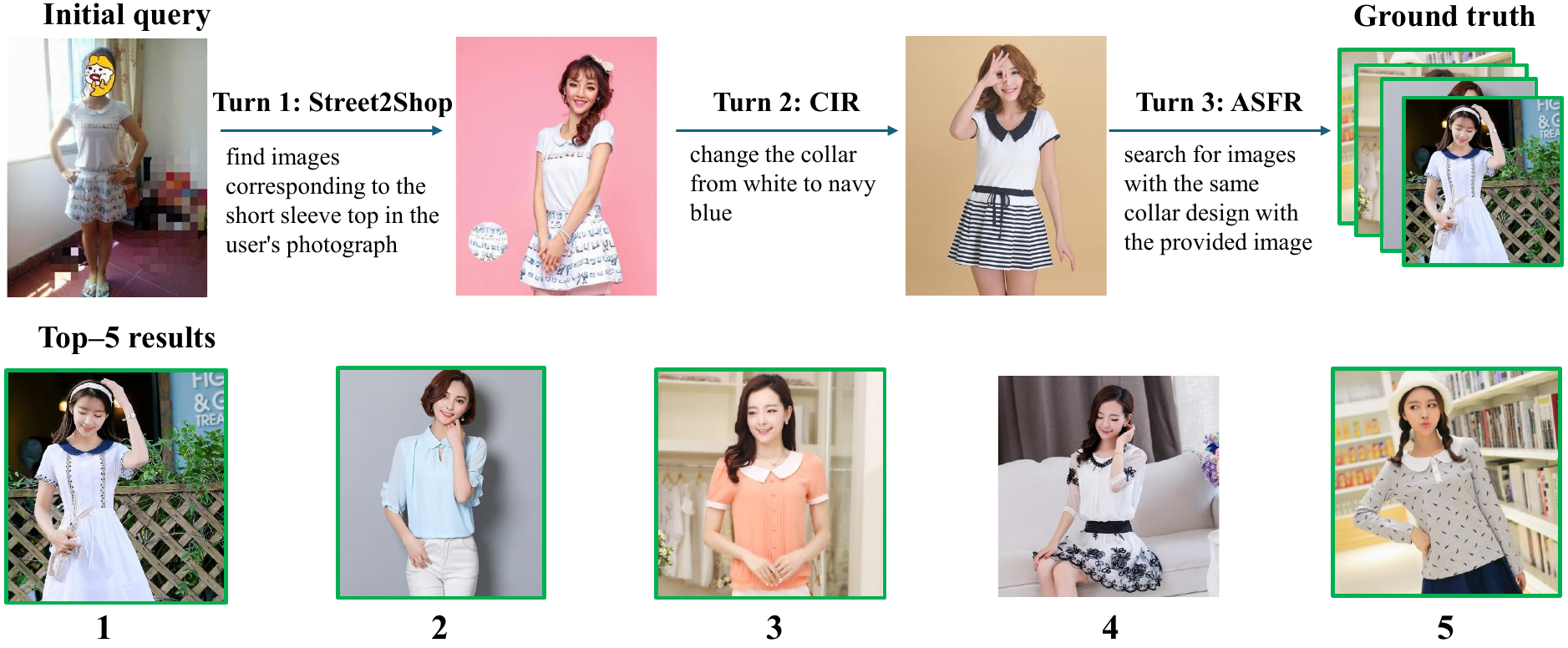}
  \end{minipage}%
}\par\vspace{0.1em}
\noindent\makebox[\textwidth][c]{%
  \makebox[0.04\textwidth][r]{}\hspace{0.01\textwidth}%
  \makebox[0.87\textwidth][l]{%
    \leaders\hbox{\rule{0.65em}{0.35pt}\hspace{0.35em}}\hfill}%
}\par\vspace{0.25em}
\noindent\makebox[\textwidth][c]{%
  \begin{minipage}[c]{0.04\textwidth}
    \raggedleft\textbf{(b)}
  \end{minipage}\hspace{0.01\textwidth}%
  \begin{minipage}[c]{0.87\textwidth}
    \centering\includegraphics[page=2,width=\linewidth]{case-study.pdf}
  \end{minipage}%
}\par\vspace{0.1em}
\noindent\makebox[\textwidth][c]{%
  \makebox[0.04\textwidth][r]{}\hspace{0.01\textwidth}%
  \makebox[0.87\textwidth][l]{%
    \leaders\hbox{\rule{0.65em}{0.35pt}\hspace{0.35em}}\hfill}%
}\par\vspace{0.25em}
\noindent\makebox[\textwidth][c]{%
  \begin{minipage}[c]{0.04\textwidth}
    \raggedleft\textbf{(c)}
  \end{minipage}\hspace{0.01\textwidth}%
  \begin{minipage}[c]{0.87\textwidth}
    \centering\includegraphics[page=3,width=\linewidth]{case-study.pdf}
  \end{minipage}%
}
\caption{Qualitative retrieval results on DIM-Fashion. Green boxes mark ground-truth positives. Case (a) shows multi-positive ASFR retrieval under background clutter, case (b) shows a successful multi-turn rollback, and case (c) is a compatibility failure in which the annotated target is absent from the Top-5 results.}
\label{fig:sup_case_studies}
\end{figure*}

Figure~\ref{fig:sup_case_studies} presents three representative three-turn sessions from DIM-Fashion. Panels (a)--(c), arranged from top to bottom, each show the initial query, the turn-wise requests and intermediate target images, and the Top-5 results for the final turn; green boxes indicate the annotated ground-truth images.

From Figure~\ref{fig:sup_case_studies}, we make three main observations. First, in case (a), although the target images contain substantial background clutter and considerable variations in pose, viewpoint, and overall appearance, FashionAM retrieves four positives within the Top-5 by focusing on the requested collar design. This demonstrates that the gallery encoder effectively suppresses irrelevant background information while preserving fine-grained, attribute-centric garment representations. Second, case (b) shows that after the session shifts from skirt modification to compatible-item retrieval, the system correctly rolls back to the skirt retrieved at Turn 1, ignores the intervening top, and incorporates the newly introduced constraints of denim material, blue color and front slit. The annotated target is ranked first. This highlights FashionAM’s ability to handle heterogeneous intent transitions and retrospective dependencies. Finally, case (c) illustrates a failure case under the single-positive evaluation protocol. Although the annotated compatible top is absent from the Top-5 results, the retrieved items are visually plausible matches for the black-and-white sneakers and remain consistent with the compatibility request. This reflects the inherently one-to-many nature of fashion compatibility, where multiple unannotated gallery items may be valid alternatives to the ground-truth image.

\end{document}